\DeclareUnicodeCharacter{2212}{\textendash}
\documentclass[journal,transmag]{IEEEtran}
\usepackage{hyperref,color,breqn,multirow}
\usepackage[top=0.7in, left=0.65in]{geometry}
\usepackage{xcolor}
\usepackage{graphicx}
\usepackage{amsmath,amsfonts,amssymb,amscd,bm}
\usepackage{tikz}
\usetikzlibrary{shapes,shapes.geometric,arrows,positioning,patterns,shapes.arrows}
\usepackage{tuda-pgfplots}
\usepackage{pgfplots}
\usepgfplotslibrary{groupplots,dateplot}
\usepgfplotslibrary{groupplots}
\usetikzlibrary{patterns,shapes.arrows,fit}
\pgfplotsset{compat=newest}

\makeatletter
\newcommand*{\rom}[1]{\expandafter\@slowromancap\romannumeral #1@}
\makeatother
\newcommand{\removelatexerror}{\let\@latex@error\@gobble}
\makeatother

\allowdisplaybreaks

\begin{document}
\title{Multi-Objective Optimization of Electrical Machines using a Hybrid Data-and Physics-Driven Approach}
\author{\IEEEauthorblockN{Vivek Parekh\IEEEauthorrefmark{1,2},
Dominik Flore\IEEEauthorrefmark{2},
Sebastian Schöps\IEEEauthorrefmark{1}} and
Peter Theisinger\IEEEauthorrefmark{2}
\IEEEauthorblockA{\IEEEauthorrefmark{1}Computational Electromagnetics Group, TU-Darmstadt, 64289 Darmstadt, Germany, sebastian.schoeps@tu-darmstadt.de}
\IEEEauthorblockA{\IEEEauthorrefmark{2}Robert Bosch GmbH, 70442 Stuttgart, Germany, Vivek.Parekh@de.bosch.com}}

\IEEEtitleabstractindextext{%
\begin{abstract}
Magneto-static finite element (FE) simulations make numerical optimization of electrical machines very time-consuming and computationally intensive during the design stage. In this paper, we present the application of a hybrid data-and physics-driven model for numerical optimization of permanent magnet synchronous machines (PMSM). Following the data-driven supervised training, deep neural network (DNN) will act as a meta-model to characterize the electromagnetic behavior of PMSM by predicting intermediate FE measures. These intermediate measures are then post-processed with various physical models to compute the required key performance indicators (KPIs), e.g., torque, shaft power, and material costs. We perform multi-objective optimization with both classical FE and a hybrid approach using a nature-inspired evolutionary algorithm. We show quantitatively that the hybrid approach maintains the quality of Pareto results better or close to conventional FE simulation-based optimization while being computationally very cheap.

\end{abstract}

}
\maketitle
\thispagestyle{empty}
\pagestyle{empty}
\section{Introduction}
\IEEEPARstart{O}{ver} the past few years, electrification in the automotive sector has contributed to reduce carbon emissions, which helps in the fight against climate change. The advancement of electrical machines is central to the electrification process. To reduce overall manufacturing costs, virtual prototyping via computer simulations, such as magneto-static finite element (FE) analysis, allows for numerical optimization of the electrical machine design before the actual machine is built. The results of the time-consuming FE analysis are further post-processed to determine various key performance indicators (KPIs), e.g., torque and power. Recent work \cite{9745918,9740189} shows how  various machine learning algorithms are used for electromagnetic device optimization. Data-driven deep learning accelerates  numerical optimization of electrical machines, as shown in \cite{9366983,9771462}. In \cite{9333549}, it is shown that different cross-domain KPIs can be predicted directly using a data-driven supervised learning approach for different representations of permanent magnet synchronous machine (PMSM), namely, parameter-based and image-based models. We presented a hybrid data- and physics-driven approach in \cite{9900437}, demonstrating that learning intermediate FE measures (electromagnetic flux, torque and non-linear iron losses) with a multi-branch deep neural network (DNN) is better than predicting direct KPIs in terms of accuracy and flexibility. Independent physical post- processing enables the fulfillment of required physical laws and the computation of complex performance curves, e.g., maximum torque limit curve. The direct approach \cite{9333549}, on the other hand, requires less computational effort and exhibits monotonous error convergence as training data increases.
\par In this contribution, we present the application of the hybrid approach \cite{9900437} for multi-objective optimization (MOO) of PMSM in a high-dimensional design space. We show how the hybrid approach aided optimization accelerates the computational process when compared to conventional FE-based workflow. This also enables possibility of finding better optimum region by doing more exploration in the large design space with a greater number of design evaluations than the conventional workflow without increasing computing effort. We perform MOO using a commercially implemented nature-inspired evolutionary algorithm \cite{optiSLangalgorithm} for contrasting target objectives, e.g., maximum power and material cost.

\begin{figure}
 	\centering
 	\input{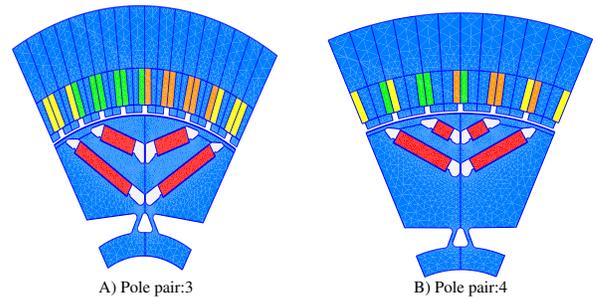}
 	\vspace{-1mm}
    \caption{Illustrations of the Double V PMSM.}
    \label{fig:RTOP}
\end{figure}

\section{Model, dataset and learning}
The double-V PMSM topology is used in this research. The procedure for data generation and the multi-branch DNN training, architecture, performance evaluation are explained in \cite{9900437}. However, a more challenging dataset has been created for, e.g., varying pole pairs, winding schemes; see dataset representative samples in \autoref{fig:RTOP}. A classical 2D magneto-static FE simulation is used to simulate the PMSM; see \cite{Salon_1995aa}. We generate $N_{\mathrm{LHS}}$ machine designs using a Latin hypercube sampling (LHS). Each machine design can be parameterized in the $i=1,\ldots,N_{\mathrm{p}}-$dimensional design space as $\mathbf{p}^{(i)}\in\mathbb{P}\subset\mathbb{R}^{N_{\mathrm{p}}}$. 
We train a multi-branch DNN using supervised learning to predict intermediate FE measures for new designs. Subsequently, desired KPIs are calculated using these predicted FE measures during the physical post-processing.  
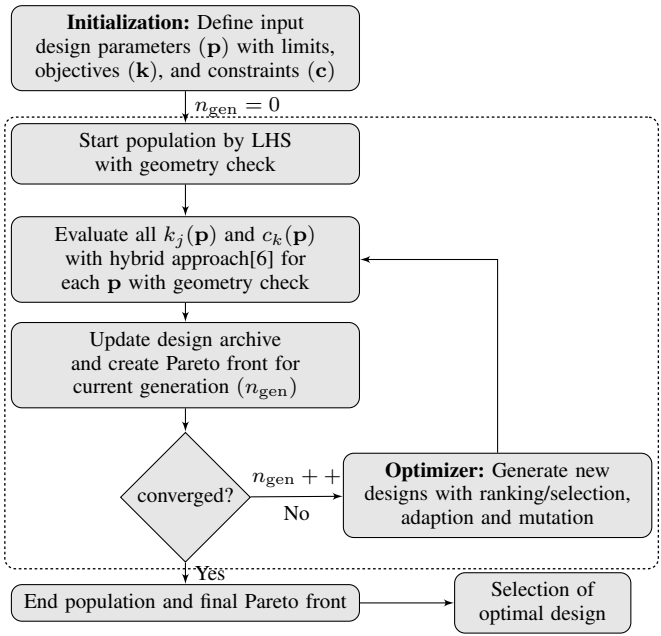
\begin{figure}
 	\centering
 	        \tikzstyle{block_round_rectangle} = [rectangle, draw, fill=black!10, 
            text width=12.5em, text centered, rounded corners,
            minimum height=1em]
            
         \tikzstyle{block_round_rectangle1} = [rectangle, draw, fill=black!10, 
            text width=11em, text centered, rounded corners,
            minimum height=1em]
            
        \tikzstyle{block_small} = [rectangle, draw, fill=black!10, 
            text width=6em, text centered, rounded corners,
            minimum height=1em]

        \tikzstyle{block_diamond} = [diamond, draw, fill=black!10, 
            text width=6em, text centered,inner sep=-1ex, minimum height=0em]
            
        \tikzstyle{line} = [draw, -latex']
        
        \tikzstyle{dotted_block} = [draw=black!, line width=0.5pt, dash pattern=on 1pt off 1pt,inner ysep=0.8mm,inner xsep=0.8mm, rectangle, rounded corners]
        
        \begin{tikzpicture}[node distance=4em]
            \tikzset{font=\footnotesize}
            \node [block_round_rectangle] (init) {\textbf{Initialization:} Define input design parameters $(\mathbf{p})$ with limits, objectives $(\mathbf{k})$, and constraints $(\mathbf{c})$};
            \node [block_round_rectangle, below of=init] (lhs) {Start population 
            by LHS with geometry check};
            \node [block_round_rectangle, below of=lhs] (maxwell) {Evaluate all ${k}_j(\mathbf{p})$ and ${c}_k(\mathbf{p})$ with hybrid approach\cite{9900437} for each $\mathbf{p}$ with geometry check};
            \node [block_round_rectangle, below of=maxwell] (pareto) {Update design archive and create Pareto front for current generation $(n_{\mathrm{gen}})$};
            \node [block_diamond, below of=pareto,yshift=-1em] (conv) {converged?};
            \node [block_round_rectangle, below of=conv] (evolutionary) {End population and final Pareto front};
            
            \begin{scope}[node distance= 1.4cm]
            \node [block_round_rectangle1, right of=evolutionary] at (2.75,-5.95) (system) {\textbf{Optimizer:} Generate new designs with ranking/selection, adaption and mutation};
            \path [line] (conv) --node[anchor=north]{No} node[anchor=south]{$n_{\mathrm{gen}}++$} (system);
        	\end{scope}

            \begin{scope}[node distance= 4.75cm]
            \node [block_small, right of=evolutionary]  (final design) {Selection of optimal design};
            \path [line] (evolutionary) -- (final design);

        	\end{scope}

            \path [line] (init) --node[anchor=west] {$n_{\mathrm{gen}}=0$} (lhs);
            \path [line] (lhs) -- (maxwell);
            \path [line] (maxwell) -- (pareto);
            \path [line] (pareto) -- (conv);
            \path [line] (conv) -- node[anchor=west] {Yes}(evolutionary);
            \path [line] (system.north) |- (maxwell.east);
            \node [dotted_block, fit = (lhs) (maxwell) (pareto) (conv)(system)] (de) {};

        \end{tikzpicture}
 	\vspace{-1mm}
    \caption{Proposed MOO workflow using hybrid approach}
    \label{fig:OPt}
\end{figure}
Design optimization of PMSM includes many conflicting objectives, constraints, and a large number of design variables. This directs to the formulation of the multi-objective optimization (MOO) problem 
\begin{align}
\label{eq:opt1}
\min_{\mathbf{p}}\quad & {k}_{j}(\mathbf{p}), & {j}=1,\dots,N_{\mathrm{k}}\\
\label{eq:opt2}
\text{s.t.}\quad & {c}_{k}(\mathbf{p}) \leq 0, & k=1,\dots,N_{\mathrm{c}}\\
&p_{l}^{L} \leq p_{l} \leq p_{l}^{U},\quad & l=1,\dots,N_{\mathrm{p}}\label{eq:opt3}
\end{align}
where ${k}_{j}(\mathbf{p})$ define objectives, ${c}_{k}(\mathbf{p})$ denote constraints which may include additional performance criteria,  ${p}_{l}^{L}$ and ${p}_{l}^{U}$ are parameter bounds. We can solve (\ref{eq:opt1}-\ref{eq:opt3}) by any common multi-objective optimizer. \autoref{fig:OPt} explains the proposed MOO-workflow, in which we employ a hybrid approach for evaluating the objectives ${k}_{j}$ and constraints $c_{k}$.
\section{Numerical results}
\begin{figure}
        \centering
        \input{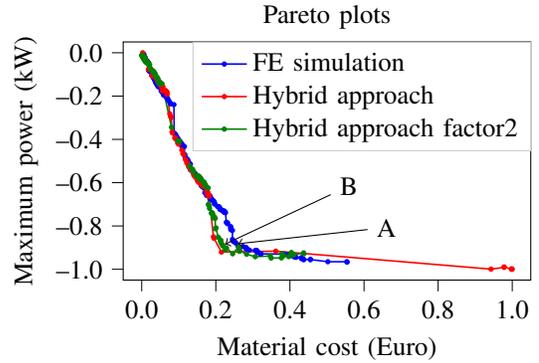}
        \vspace{-3.5mm}
        \caption{Pareto-fronts for Material cost and Maximum power}
        \label{fig:moo}
\end{figure}

\begin{figure}
    \centering 
 	\input{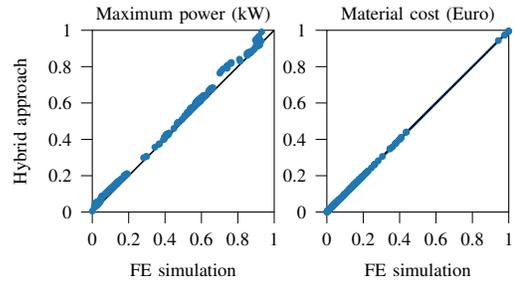}
    \vspace{-3.25mm}
    \caption{Prediction plot of Pareto designs for hybrid approach}\label{fig:kpi_pred}
    \label{fig:eval}
\end{figure}

The MOO with the nature-inspired evolutionary algorithm \cite{optiSLangalgorithm} is demonstrated for two competing KPIs, i.e., maximum power and material cost. We run the MOO workflow in parallel with both the classical and hybrid approaches, with identical settings (e.g., design selection criteria, mutation rate, convergence settings, ranking method, etc.) and initial designs. The results show that the final Pareto-front for the hybrid approach is close to (or even better than) the one from the FE-based workflow. The hybrid approach reduces the overall computational cost by about eight times. Therefore we can afford to run the hybrid approach with twice as many evaluations. We call this `factor2' in \autoref{fig:moo}. This is still four times faster than the FE-based approach and allows to explore the optimal region more precisely. \autoref{fig:eval} depicts the prediction plot for two KPIs for all the feasible samples predicted in the hybrid Pareto fronts. Finally, \autoref{fig:RTOP} presents designs A and B from the Pareto-fronts of the FE simulation and hybrid approach, respectively. These designs are close from the evaluation point of view of the maximum power objective.

\section{Conclusion}
This paper presents how the hybrid approach is used in the MOO cycle to replace the FE-based solver. The numerical results show that the proposed hybrid-approach-based MOO workflow is close to the FE-based workflow at a significantly lower computational cost and with reasonable numerical accuracy. This also enables efficient exploration of design space to explore the optimal region more precisely while being computationally cheap. This implies that more novel design generation is possible in the large complex input design space.

\bibliographystyle{ieeetran}
\bibliography{main_ref}

\end{document}